\documentclass[letterpaper, 10 pt, conference]{ieeeconf}  

\IEEEoverridecommandlockouts                              
\usepackage[utf8]{inputenc}
\usepackage[T1]{fontenc}
\usepackage{float}
\usepackage[table,xcdraw]{xcolor}
\usepackage{graphicx}
\usepackage{hyperref}

\title{\LARGE \bf
Data-driven behavioural biometrics for continuous and adaptive user verification using Smartphone and
Smartwatch
}

\author{Akriti Verma$^{1}$, Valeh Moghaddam$^{2}$ and Adnan Anwar$^{3}$
\thanks{*This work was not supported by any organization}
\thanks{$^{1}$School of IT, Deakin University
        {\tt\small vermaakr@deakin.edu.au}}%
\thanks{$^{2}$School of IT, Deakin University
        {\tt\small valeh.moghaddam@deakin.edu.au}}%
\thanks{$^{3}$School of IT, Deakin University
        {\tt\small adnan.anwar@deakin.edu.au}}%
}

\begin{document}
\pagenumbering{arabic}

\maketitle
\thispagestyle{plain}
\pagestyle{plain}

\begin{abstract}

Recent studies have shown how motion-based biometrics can be used as a form of user authentication and identification without requiring any human cooperation. This category of behavioural biometrics deals with the features we learn in our life as a result of our interaction with the environment and nature. This modality is related to change in human behaviour over time. The developments in these methods aim to amplify continuous authentication such as biometrics to protect their privacy on user devices. Various Continuous Authentication (CA) systems have been proposed in the literature. They represent a new generation of security mechanisms that continuously monitor user behaviour and use this as the basis to re-authenticate them periodically throughout a login session. However, these methods usually constitute a single classification model which is used to identify or verify a user. This work proposes an algorithm to blend behavioural biometrics with multi-factor authentication (MFA) by introducing a two-step user verification algorithm that verifies the user's identity using motion-based biometrics and complements the multi-factor authentication, thus making it more secure and flexible. This two-step user verification algorithm is also immune to adversarial attacks, based on our experimental results which show how the rate of misclassification drops while using this model with adversarial data.

\end{abstract}

\section{Introduction}

Maintaining the security of digital and non-digital assets requires the capacity to identify or authenticate a person. Passwords or physical tokens (such as ID cards) are frequently used to offer security, although they are easily stolen or reproduced. Biometric approaches, which are based on a person's unique physical or behavioural features, do not suffer from these drawbacks \cite{jain2011introduction}. The fingerprints or iris of a person are used in most physical biometric systems. Such methods can be difficult to use at times, and behavioural biometrics offers a viable alternative. Smartphones and smartwatches, which are equipped with powerful sensors, provide a convenient platform for implementing and deploying mobile motion-based behavioural biometrics. Popular mobile devices, such as smartphones and smartwatches, feature motion sensors that can be used as the foundation of a biometric system, making motion-based biometrics an especially appealing option. Biometrics based on these devices can be employed as a main authentication and identification technique or as part of a multi-factor system. It was recently studied that the accelerometer and gyroscope sensor on both the smartphone and smartwatch, can be used in combination to perform motion-based behavioural biometrics \cite{weiss2019smartphone}. 

It has already been realised that authentication with just a single factor is not reliable enough to provide adequate protection due to many security threats. Multi-factor authentication mechanisms are thus required to enforce strong authentication based on the biometric and identifiers of other nature \cite{dasgupta2017multi}. Multi-Factor Authentication (MFA) is a secure process of authentication that requires more than one authentication technique chosen from independent categories of credentials. For the most part, MFA is based on biometrics, which is the automated recognition of individuals based on their behavioural and biological characteristics. This step offers an improved level of security as the users are required to present evidence of their identity, which relies on two or more different factors  \cite{bhargav2007privacy}, \cite{ometov2018multi}.

Since motion-based biometrics rely heavily on user data and machine learning models, it is essential to ensure that these models cannot be easily corrupted and therefore this work also analyses the impact of adversarial attacks on the developed continuous and adaptive user verification algorithm by performing a Zero Order Optimisation attack on user data and then formulating a strategy to defend the model against such attacks.
\subsection{Aim \& Objectives}~\label{subsec:aims}
This paper aims to push the ongoing research towards introducing behavioural biometrics for user verification by proposing a user verification model that can be integrated with the current multi-factor authentication systems. It also examines the risks associated with using motion-based biometrics by performing an adversarial attack and formulating a defence strategy to defend the model against the attacks.

The successful completion of this work would make the following research contributions:
\begin{itemize}
    \item This work leverages the relative value of smartphone and smartwatch based accelerometers and gyro sensors for motion-based biometrics. It presents a continuous and adaptive user verification algorithm that can be integrated with multi-factor authentication.
    \item It analyses the proposed continuous and adaptive user verification algorithm for its vulnerability towards adversarial attacks.
    \item It also provides a strategy to defend the continuous and adaptive user verification algorithm against adversarial attacks as well as future prospects leading to the safe integration of motion-based biometrics for user verification.

\end{itemize}
\subsection{Structure}~\label{subsec:structure}
The rest of this paper is structured as follows. We start by presenting a literature review of motion-based biometrics before describing the topic's background and related work. Then we put forward the problem statement which is followed by a brief on the proposed algorithm and explain the proposed user verification approach along with the design and methods utilised in this work and empirical analysis of its implementation and discuss its vulnerability towards adversarial attacks before summing up a conclusion.  


\section{Literature review}

In recent years the user authentication model is gradually shifting from "something the user knows" to "something the user is". Behavioral biometrics (BB) and continuous authentication (CA) are used in this method \cite{stylios2016review}, \cite{stylios2021behavioral}. As mobile devices grow increasingly technologically capable, it is evident that the built-in sensors can be utilised to effectively capture most users' activity, enabling behavioural biometric user authentication. By continuously monitoring user behaviour and re-authenticating user identification throughout a session, CA technology adds an extra layer of protection to the original login procedure \cite{crouse2015continuous}, \cite{shi2011senguard}. Behavioural biometrics authentication approaches are based on a person's behavioural characteristics such as walking stride, touch gestures, keystroke dynamics, behaviour profiling, hand waving, and power consumption and fusion. However, due to some key flaws, such as the possibility of false positives/negatives, the balance between security and usability, privacy concerns, and so on, behavioural biometrics authentication is limited. To address these flaws, it's vital to increase accuracy and look into how to strike a balance between security and usability. Because users' habits and behaviour may vary over time, authentication systems must be able to adapt to these changes. It has also been studied that adversarial examples can be crafted for machine learning and deep learning models which when fed to these models result in high misclassification errors \cite{benegui2020adversarial}. Therefore, in this paper, we present a two-step motion-based user authentication model, which although relies on a pre-trained model but also verifies the user each time. Since user verification is treated as a classification problem in this work, we also assess the model's vulnerability to adversarial examples and propose a defence strategy.

Although there are a great number of studies proposing several different methods of identifying and verifying users based on their device interactions and the range of motion-based activities recorded by the sensors in their devices, the field of Continuous Authentication (CA) using behavioural biometrics needs a strategy to evaluate its efficiency and a framework to implement the theoretical background that has been developed as a result of vast research in the field. A viable framework for employing behavioural biometrics as a means for user authentication will not only boost a high degree of confidence in the field but also help produce results that match the expectations and requirements of authentication systems. Most of the observations made with behavioural biometrics support the hypothesis that a person's behavioural profile can be developed by utilising their device usage/interaction patterns. However, this has also been a consistent point of scepticism as human behaviour tends to change or evolve with time. This work also talks about the prospect of how behavioural biometrics can adapt to the changes in human behaviour with time.

It is also important to consider the risks which come with using behaviour profiles and motion-based biometrics, apart from the data collection and its safe and ethical usage, human behaviour can be imitated. Stylios et. al. \cite{stylios2021behavioral}, \cite{muaaz2017smartphone} talks about the various kinds of zero-effort attacks that can be planted on a behavioural biometrics system. The attacker does not take any complex action in the zero-effort attack. It is predicated on the attacker's and legitimate user's templates being sufficiently similar, and it is related to the `uniqueness' property of a biometric trait. An adversary attack entails the attacker doing specific actions in order to convincingly mimic a real user. The amount of sophistication of an adversary attack is highly dependent on the resources available, such as digital or physical means, time, and information about the biometric system and the victim. This reinforces the importance of real-time monitoring and evaluation of the results produced by the underlying machine learning/deep learning models in these methods. This work also dives into analysing the vulnerability of the proposed user verification model by testing it against an adversarial attack. One way of minimising the risk and enhancing the reliability of behavioural biometrics is to combine the authentication or verification process with other approaches to confirming the user's identity. Authentication accuracy is improved by combining behavioural biometrics with a password or token-based authentication, which has been intensively researched in the literature. A plethora of studies has emphasized the superiority of multimodal biometric approaches to single biometric methods \cite{shi2010implicit}, \cite{eshwarappa2011multimodal}, \cite{neha2016review}.


\section{Background \& related work}

Biometrics-based identification is the next frontier in user verification and authentication since it is more efficient than digital passwords/PINs or cryptography-based digital signatures. Biometrics are not susceptible to theft and cannot be misplaced or forgotten. They recognise a person based on their distinct traits and hence provide a unique verification for each user. Biometrics-based identification and authentication can be easily confirmed by using the underlying machine learning or deep learning model alone or in combination with a multi-factor user authentication system. Aside from these advantages, behavioural biometrics are favoured over pattern matching-based biometrics like EEG/ECG matching, iris detection, or palm-veins matching because these require specialised apparatus that might be costly and inconvenient. Behavioural biometrics are based on sensors available in our smartphones and smartwatches and work to discover a pattern based on a user's daily actions. They may not even require human input to identify a person. A strong user authentication system requires some key elements for a reliable authentication, such as an aspect of 'knowledge' (something only the user knows; for example, length or complexity), an aspect of 'possession' (something only the user possesses; for example, a characteristic movement), as well as an aspect of 'inherence' (something only the user possesses; for example, a characteristic movement) (something the user is; a fingerprint or biometric specification). Using behavioural biometrics not only fulfils all of these categories of a user's credentials but also considerably reduces the chance of these authentication aspects being compromised. Therefore, behavioural biometrics permits enhancing the user experience as user authentication/identification can be transparent for them.

\subsection{Behavioural biometrics }
Behavioural biometrics-based user identification or verification based on a single activity has been detailed in various research. S. R. Sudhakar et al. \cite{sudhakar2021actid} offer one such framework for user identification, in which they show how a person's hand motions while walking may be recognised using the accelerometer and gyrosensor in their smartwatch. For the goal of identifying a user, the framework uses a correlation-based feature evaluation and selection method, as well as a sliding window-based voting classifier. As a result, it meets a number of key design requirements for gait authentication on resource-constrained devices, such as lightweight and real-time classification, high identification accuracy, and a small number of sensors. However, it only focuses on GAIT-based identifying activities.

R. Oak et al. offer another architecture for continuous authentication using behavioural biometrics, in which they suggest a novel approach for authentication based on the concept of a logical DNA that integrates many factors to generate a user profile. Using machine learning models such as k-Nearest Neighbors, Random Forests classifiers, and a 1D Convolutional Neural Network \cite{8455040}. R. Luca et al explain how inertial data from accelerometers may be utilised to authenticate a user by identifying users using data from walking and computational activities \cite{9497421}. Su et al. offer a strategy for GAIT identification that can gradually merge temporal features while extracting spatial features to achieve spatiotemporal feature extraction. By forwarding partial channels of feature maps and fusing features from consecutive frames, the model extracts temporal information, and it modifies the part-based method to split the feature map into numerous parts, which refines the spatial features \cite{9506490}. As can be seen, GAIT is the focus of the majority of behavioural biometrics research.

\subsection{Adversarial attacks for machine learning based biometrics}
Malicious applications and attacks on user devices and data have become common. An attacker's main goal is to take control of mobile devices that are protected by authentication systems in order to obtain access to a user's private information or to perform non-permitted operations. Adversarial examples are data samples with minor modifications that, when fed into machine learning or deep learning models, result in inaccurate predictions. Different approaches to creating adversarial instances can have different outcomes. The purpose of creating an adversarial example from a real one is to induce a misclassification error, i.e., the adversarial attack is about making erroneous predictions \cite{wang2020user}. White-box and black-box attacks on machine learning classifier models are the two types of adversarial approaches. The attacker is assumed to have complete access to a fully differentiable target classifier in white-box attacks (weights, architecture, and feature spaces) whereas in the case of black-box attacks, without access to the target models to compute gradients, the adversary attempts to generate malicious perturbations. However, it is unrealistic to assume that the adversary has access to the authentication model when considering adversarial attacks in the domain of machine learning or deep learning-based authentication \cite{tan2019adversarial}. For example, in a highly secure setting, the model could be located on a distant server, away from the target machine. As a result, the attacker would be unable to undertake the black-box attacks, being recommended by the authors in \cite{papernot2017practical} because they would be unaware of the outcome of the authentication computed by this remote server and the black-box adversary can only observe outputs given by the model to chosen inputs. White-box attacks that rely on gaining access to the model's architecture and weights are also unfeasible. Therefore, for this experiment, a black-box attack has been chosen, which relies on the data to create adversarial samples and does not require any information about the underlying model. 

Although knowledge-based user authentication has been the most popular means of validating the identity of the users for a long time, a fusion of basic physiological and behavioural biometrics have attracted a lot of research lately as a medium to reduce the vulnerability of knowledge-based authentication. When used alone, physiological biometrics need expensive equipment and are therefore not the most widespread, whereas behavioural biometrics have low accuracies when used on their own. By fusing one or more of these techniques according to efficiency and availability, multi-factor authentication can assist overcome the limitations of these approaches and hence enhance the security of the authentication process. Furthermore, by combining behavioural biometrics, the amount of user involvement required for each session when the user's identification must be validated can be reduced. According to C. Wang et al., the trend in mobile device authentication is multi-factor authentication, which determines a user's identification by integrating (rather than just combining) multiple authentication metrics. For example, when the user inputs the knowledge-based secrets (e.g., PIN), the user's behaviour biometrics (e.g., keystroke dynamics) could be extracted simultaneously, providing enhanced authentication while sparing the user the trouble of conducting multiple inputs for different authentication metrics \cite{wang2020user}.


\section{Problem description}
Research on the performance of behavioural biometrics with daily activities for both users and identification of activities has been extensive and can thus be used to identify users or their activities. But today the authentication systems are multi-step, continuous and adaptive as the diversity of device use develops daily. This combination was developed to improve security and to provide an easy and smooth activity session of the user device interactions. It has also been studied that various forms of authentication can be used as an aggregate to validate the identity of the users. This research is built on the same idea of using different forms of user identification to match various elements of a secure authentication system (knowledge, possession and inherence), and so introducing behavioural biometrics into existing multi-factor authentication systems. In this work we outline how behavioural biometrics be modelled and readily incorporated into present-day user authentication systems to enhance their security by utilising the theoretical observations and results obtained from research in this area.

Apart from the high accuracies of user and activity recognition using behavioural biometrics, research in this field has also shown how the machine learning and deep learning models behind these identification or verification algorithms are highly susceptible to various risks and attacks. These attacks being simple in nature, result in low confidence towards employing these algorithms for user authentication. It is, therefore, essential that the verification algorithms are analysed for their exposure to such attacks. This paper proposes an algorithm to integrate behavioural biometrics as a part of the multi-factor authentication process thus paving a way for the evolution of motion-based biometrics and future research in this area. Not only do we formulate a way to identify or verify a user but also analyse the risks associated with implementing this process of user verification (which largely entails the use of machine learning models) in a real-time environment and develop a strategy based on the model's prediction score to mitigate this risk by empirically evaluating if the model is safe to be trusted or not. Through this algorithm, we propose how behavioural biometrics can be implemented as a means of user verification while minimising the risk of misclassification when identification and authentication tasks are modelled as classification problems.
\begin{table}[!htbp]
\centering
\label{table:activities}
\caption{The physical activities}
\begin{tabular}{cc}
\hline
\multicolumn{2}{c}{\textbf{Non-hand oriented activities}}    \\ \hline
\textbf{A}          & walking                                \\
\textbf{B}          & jogging        \\
 
\textbf{C}          & stairs                                 \\
\textbf{D}          & sitting                                \\
\textbf{E}          & standing                               \\
M                   & kicking a ball                         \\ \hline
\multicolumn{2}{c}{\textbf{Hand-oriented activities}}        \\ \hline
\textbf{F}          & typing                                 \\
\textbf{G}          & teeth                                  \\
\textbf{O}          & catch                                  \\
\textbf{P}          & dribbling                              \\
\textbf{Q}          & writing                                \\
\textbf{R}          & clapping                               \\
\textbf{S}          & folding                                \\ \hline
\multicolumn{2}{c}{\textbf{Hand-oriented eating activities}} \\ \hline
\textbf{H}          & soup                                   \\
\textbf{I}          & chips                                  \\
\textbf{J}          & pasta                                  \\
\textbf{K}          & drinking                               \\
\textbf{L}          & sandwich                              
\end{tabular}
\end{table}
Continuous authentication reviews the legitimacy of a user during each session, thus reducing the risk of session hijacking \cite{9530399}. These authentication techniques are aimed mainly at the ability to authenticate a user's identification conveniently and reliably and to continually reconfirm the user's identity. Integrating behavioural biometrics with multi-factor authentication will provide a secure ground for their foundation in real-time authentication systems and will motivate further research in this area. The idea behind multi-factor authentication is to use multiple ways to confirm the user's identity. This work enables the use of behavioural biometrics as a part of the multi-step authentication process. It has been studied how motion-based biometrics from accelerometers and gyrosensors are easy and no-interaction based techniques to validate the identity of a user and this study uses the transformed WISDM dataset which contains user records sampled at a rate of 10 seconds and thus implies that the process of authentication can be based on a short sample of data from the user such as 10 seconds. We present how motion-based biometrics can be used to improve the ease and reliability of multi-factor authentication process.


\section{Proposed user verification model}

This work utilises the recently published WISDM dataset (publicly available) and its experimental results based on four sensors (accelerometers and gyro sensors from both smartphone and smartwatch) and 18 activities (Table-\ref{table:activities}) to develop a mechanism to implement continuous and adaptive user verification using behavioural biometrics as a part of multi-factor authentication \cite{weiss2019smartphone}.
\begin{figure}[!htbp]
        \centering
        \includegraphics[width=8cm,height=8cm,keepaspectratio]{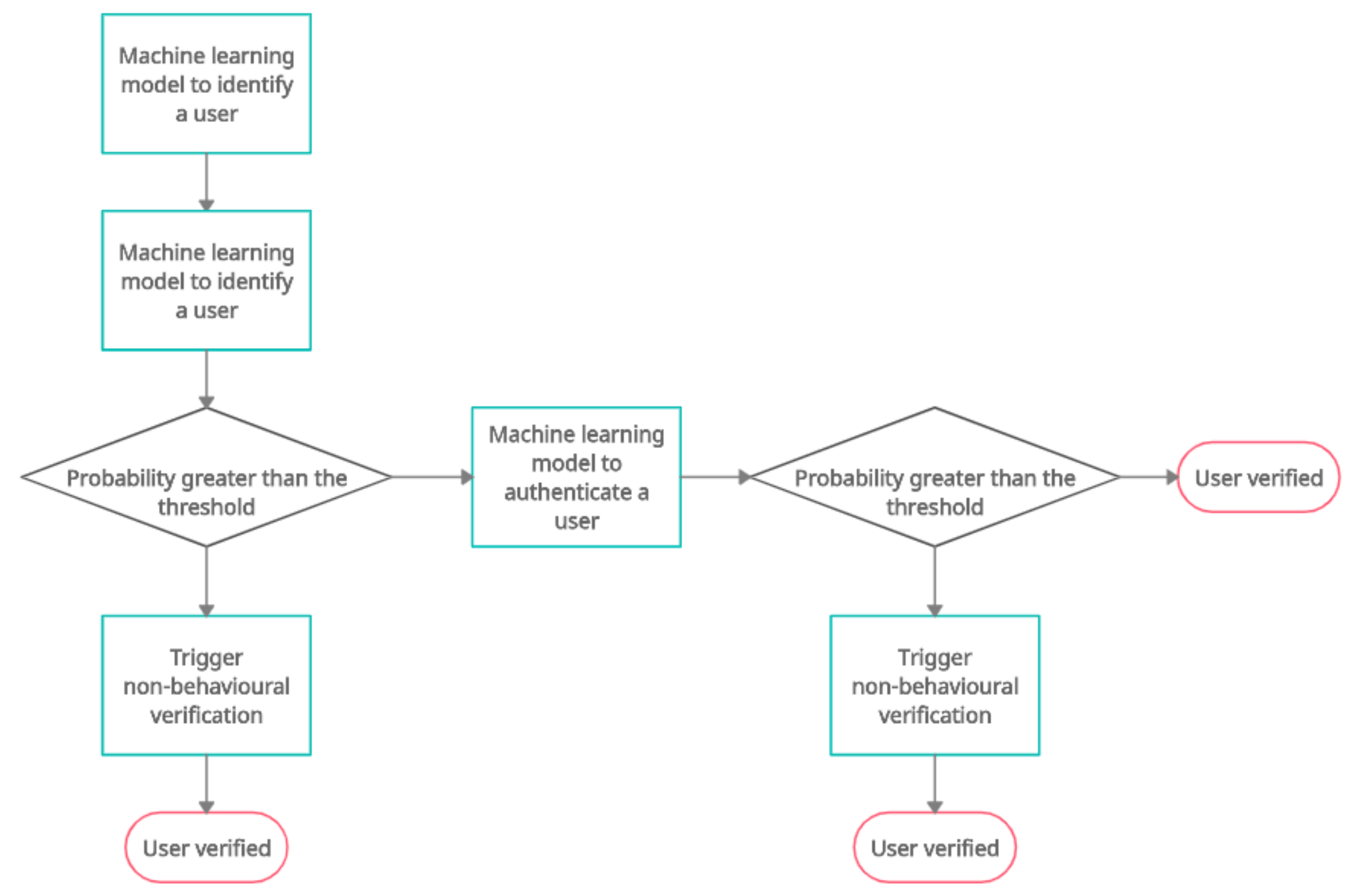}
        \caption{User verification algorithm}
        \label{fig:my_label_vr}
    \end{figure}
\begin{figure*}[!htbp]
        \centering
        \includegraphics[width=12cm,height=12cm,keepaspectratio]{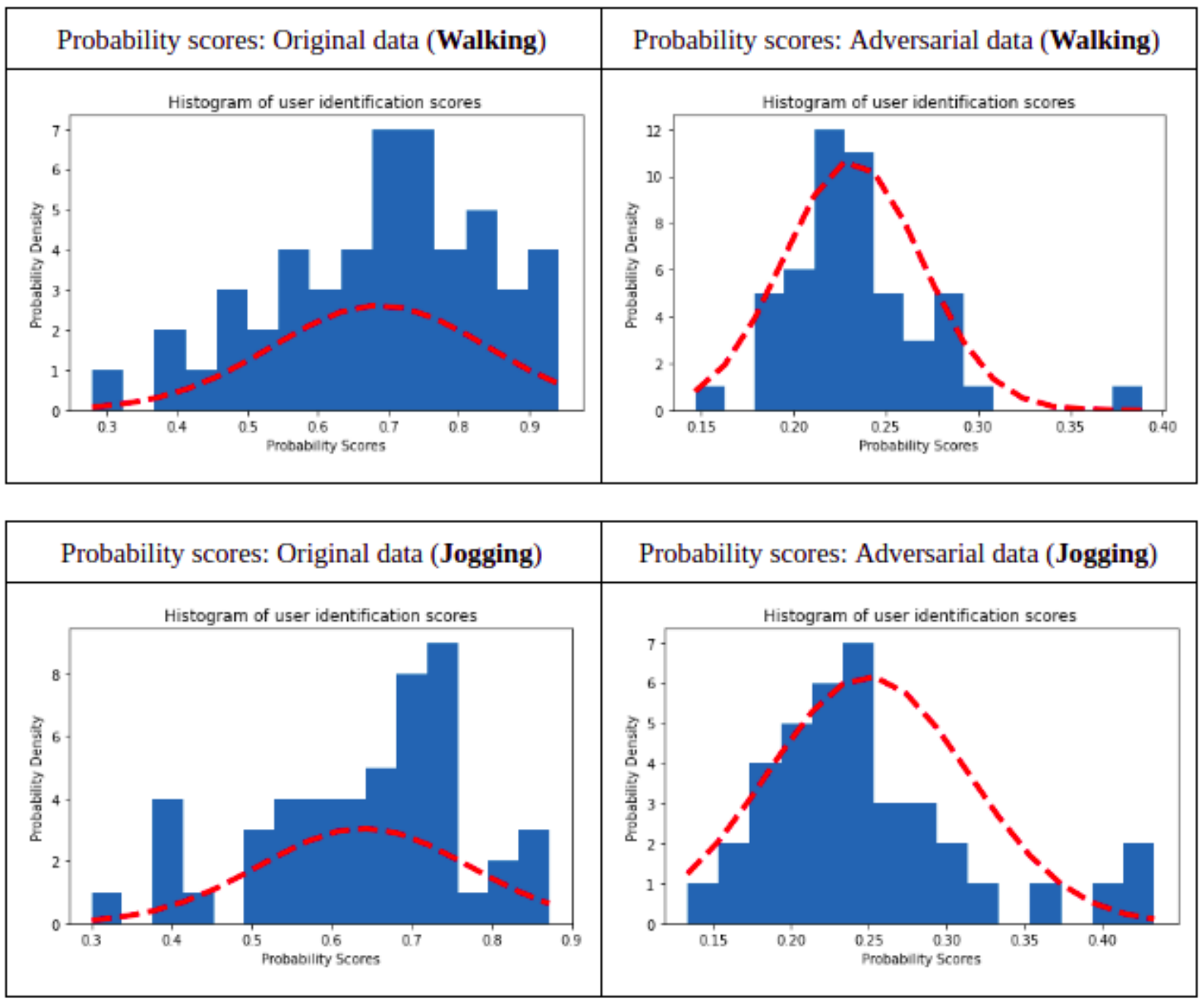}
        \caption{Prediction probabilities for Identification-Non-hand oriented activity}
        \label{fig:my_label}
    \end{figure*}
Weiss et. al. in 2019 described how simple machine learning models achieve high rates of accuracy on the task of identifying and authenticating users based on their motion biometrics using activities of daily life. They performed the experiment on the WISDM data where they categorised the recorded activities into three classes, namely, non-hand oriented activities (for example walking, jogging etc), general hand-oriented activities (for example clapping, typing etc) and hand-oriented eating activities (for example eating a sandwich, pasta etc) \cite{weiss2019smartphone}, \cite{weiss2019wisdm}. Since the WISDM dataset contains data from both the smartphone and smartwatch, they conducted the experiment on various combinations of four sensors (phone-accelerometer,  phone-gyrosensor, watch-accelerometer and watch-gyrosensor). For both the tasks, i.e. user identification and authentication, they trained a separate machine learning model for each activity. The analysis tested three different machine learning models namely Random Forest, Decision Tree and K-nearest neighbours out of which Random Forest performed the best for all activities and every sensor combination. This was one of the first studies focussing on daily life activities for the purpose of user identification or authentication.

This work plans to take up the analysis by Weiss et. al. and employ the results to build a user verification system that is ready to be integrated and utilised as a part of the current day user authentication systems. As the purpose of this algorithm is to authenticate users, it is essential that the underlying machine learning models are tuned to achieve better accuracy. In the hunt for achieving higher accuracy, various other machine learning models were employed, however, the best results from Weiss's analysis could not be beaten for each activity. Similarly, when deep learning models were trained on WISDM data, high accuracy for every activity could not be produced due to factors such as the size of the dataset (when factored for each activity).

The proposed user verification algorithm as shown in Figure-\ref{fig:my_label_vr} is a multi-step process, which can easily be translated as a part of the multi-factor authentication system. The algorithm contains two machine learning models, for identification and authentication. The identification model is different for each activity, whereas the authentication model is different for each activity as well as for every user. The idea is to identify the user and then verify the identification by confirming that it can be differentiated from an imposter. It has also been taken into consideration that the machine learning model cannot always be trusted, however developing a completely reliable model is a work in progress. As per our hypothesis, we have identified a threshold value for every model (each activity has a different model), whenever the probability score of a model's prediction i.e. identification or authentication (identification and authentication have been modelled as classification problem) is equal to or greater than the threshold value, the model can be trusted with behavioural biometrics and proceed for verification. If this is not the state, it is safer to continue using normal means like OTP or verification code to verify the user.
\begin{figure*}[]
        \centering
        \includegraphics[width=12cm,height=12cm,keepaspectratio]{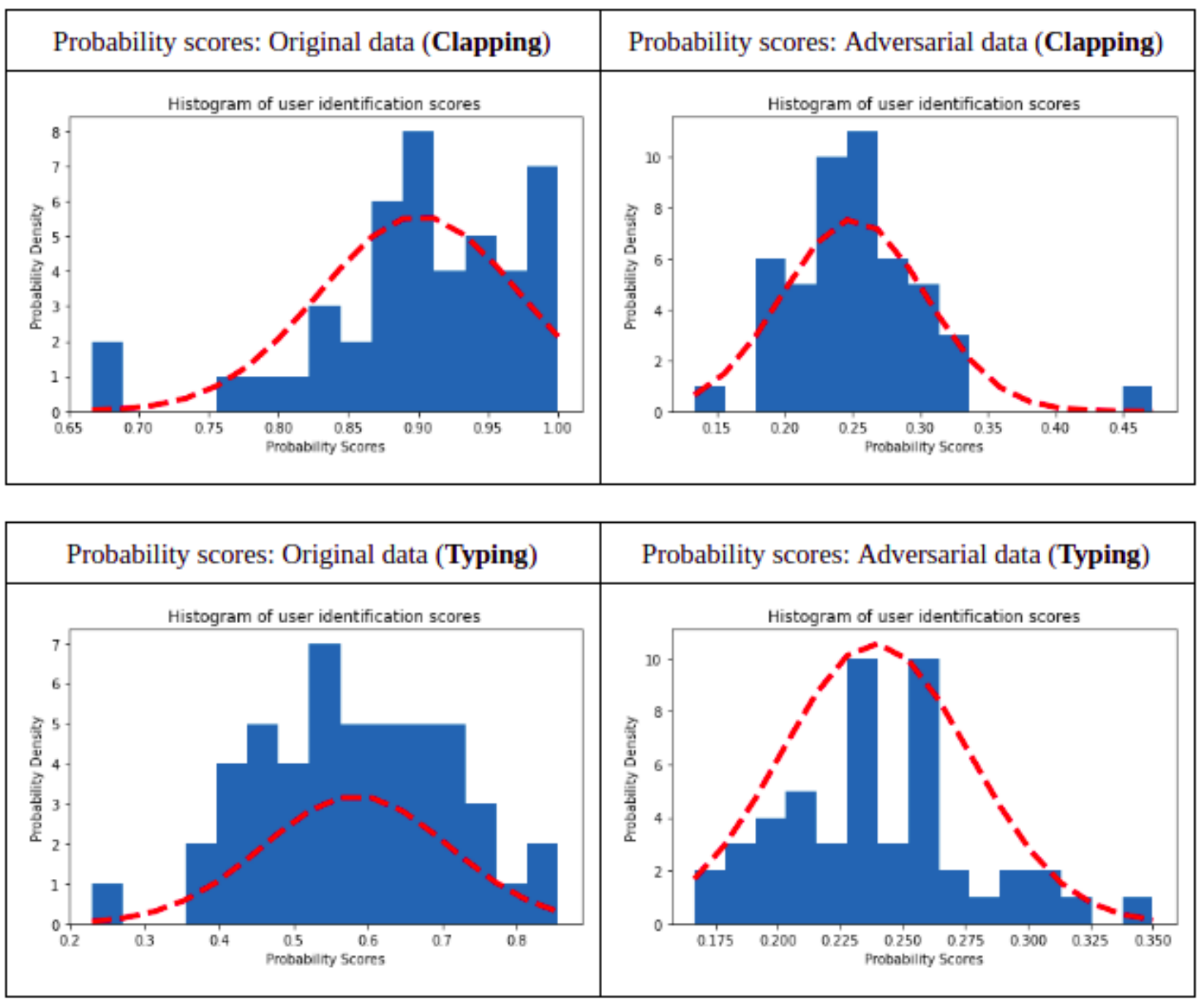}
        \caption{Prediction probabilities for Identification: General-hand oriented activity}
        \label{fig:my_label}
    \end{figure*}

However, integrating this algorithm as a part of a real-time user authentication system requires it to be secure and be able to enhance the reliability of the overall system. Although the system hosting the machine learning models may not be easily accessible by the intruder, the data used to train the models is comparatively easier to dupe, which when perturbed, will significantly impact the performance of the models. Therefore, in order to assess the vulnerability of these machine learning models to attacked or altered data, the model was tested for black-box adversarial attacks. The WISDM data was attacked with a zero-order optimisation attack and as predicted, the accuracy of the model was remarkably affected. On investigating the misclassifications further, it was found that the model was misclassifying the samples with a very low confidence score as compared to the classifications on the original data, and hence it was ruled that the model cannot be trusted with adversarial data due to its low confidence and prediction errors. Therefore, it was concluded that the confidence score can be used as a threshold to determine when the model can be trusted. Thus the proposed user verification model works by validating this threshold each time to rule out the possibility of intrusion.


\section{Experiment methodology}
The data for 51 individuals were recorded over 18 different activities of daily life (Table-\ref{table:activities}) in the WISDM dataset, which was released in 2019 \cite{weiss2019wisdm}. The data set includes low-level time-series sensor data from the phone's accelerometer, gyroscope, accelerometer, and gyroscope, as well as data from the watches' accelerometer and gyroscope. All of the time-series data is tagged with a subject identity in addition to the activity being performed, allowing the data to be used for constructing and evaluating biometrics and activity recognition models \cite{weiss2019smartphone}. 
\begin{figure*}[]
        \centering
        \includegraphics[width=12cm,height=12cm,keepaspectratio]{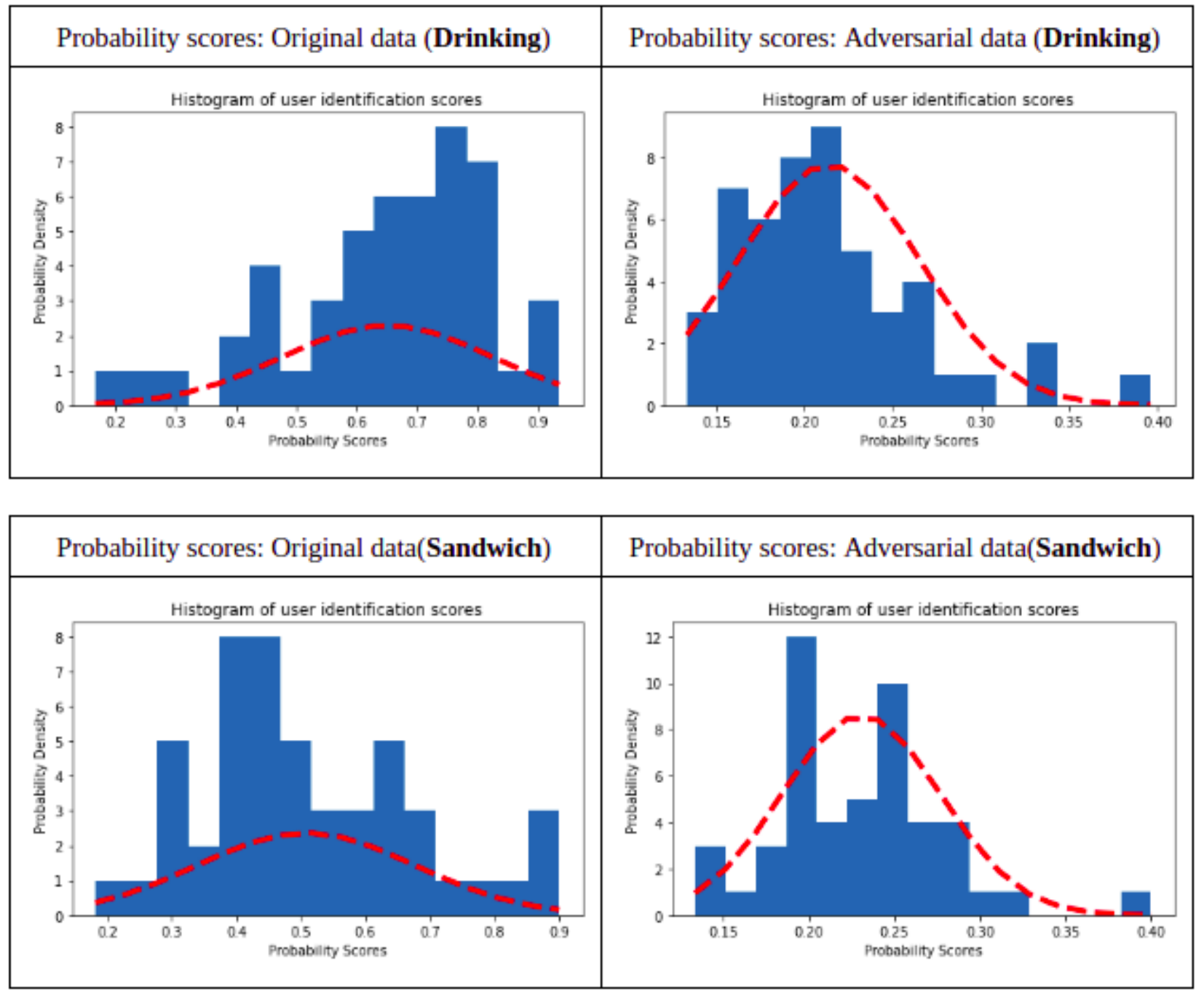}
        \caption{Prediction probabilities for Identification: Hand oriented eating activity}
        \label{fig:my_label}
    \end{figure*}

A model with very high accuracy is necessary for building an algorithm to provide continuous and adaptive user authentication, for which multiple deep learning methods such as CNN, LSTM, and densely connected neural networks were applied. However, because these models take a lot of data to train and the task is focused on authentication, data augmentation may not be the best option, the focus was turned back to machine learning models. Although algorithms such as XGBoost and SVM fared well, Random Forest still dominated when it came to average accuracy across a variety of activities as described in \cite{weiss2019smartphone}. 

The methodology used for identifying a user is as follows. As mentioned previously, both identification and authentication tasks have been treated as classification problems for this experiment. This is because this treatment is a widely accepted method for identifying and authentication users which are proven to produce results in the literature \cite{alzubaidi2016authentication}, \cite{dave2021iot}, \cite{shopon2021biometric}. Each subject represents a separate class throughout the identification experiment; there are fifty-one classes in the identification data set (since the WISDM data contains records for 51 users).In this situation, the training set must include data from all of the participants, hence the training and test sets' subjects should overlap. Stratified 10-fold cross-validation is used to partition the training and test data so that both sets have the same percentage of data from each subject. Models of identification are created for each activity. For the purpose of testing the identification, a multi-class prediction problem, the accuracy of the machine learning models was used as the metric. The model for every activity was then tested for adversarial attacks by feeding adversarially generated data (using ZOO attack) to test the models.

The experimental method for authentication was quite different from that of identification. The authentication task entails distinguishing between a legitimate subject and an intruder. As a result, authentication is a two-class classification problem. Each model is built utilising data from the subject being verified as well as data from ‘‘other" subjects that are grouped into a single class. Because data from actual imposters will not be available in real-world scenarios, it is crucial to ensure that the ‘‘imposters" in the training and test sets do not overlap. Furthermore, authentication model training data must be appropriately partitioned, as a training set with a significant degree of class imbalance will be prejudiced against authenticating a real user. The authentication models are built for every user and for every activity of a user.
 
Because the user-verification process is so reliant on the underlying machine learning models, it's critical to assess the safety and security of these models. Therefore, in order to test this model, adversarial data was generated for the RandomForest model using IBM's ART package, and the model was attacked with a Zero Order Optimisation attack, during which the model's accuracy fell from 97.6 to 28 percent. The probability scores of the model's classification output were studied as a preventive measure (user identification and authentication are treated as classification tasks), and it was discovered that when adversarial examples are introduced, the probability scores for identification drop significantly, indicating that the model is not confident in the adversarial predictions. To protect the model from such attacks, a threshold value was chosen for each model (each activity has its own model), and identification is now only trusted if the classification probability is above the threshold. This adversarial technique is a two-step procedure that can be easily used with multi-factor authentication to make the process more secure.

Since the two steps in our user verification model are different procedures, i.e. identification and authentication, this experiment for analysing the vulnerability of the machine learning model on adversarial attacks was also performed for user authentication. Although the accuracy of authentication was significantly affected, the accuracy of the model was not used as the metric for the case of authentication. The Equal Error Rate (EER), a standard metric for comparing authentication methods \cite{gafurov2006biometric}, is used to evaluate authentication performance in this work. The False Acceptance Rate (FAR), which is the rate at which the model erroneously accepts an imposter as a valid user, equals the False Rejection Rate (FRR), which is the rate at which the model incorrectly rejects a legitimate user, is used to calculate this measure. The probability threshold used to assign a classification can be changed to change FAR and FRR. When working on the user verification task, this statistical value is utilised to show biometric performance. On a ROC curve, the EER is the point where the false acceptance rate and false rejection rate are equal. In general, the smaller the equal error rate, the higher the biometric system's accuracy.

\subsection{Defending the proposed user verification model}
Continuous authentication offers a layer of protection to the service provider, which is utilised to increase usability in most circumstances \cite{choi2021keystroke}. As it is shown in the user verification model, the process of verifying the identity of the user is multi-step and can be time-consuming depending on the configuration of the underlying system. The advantage that is produced out of using this method and spending a little extra time in initially confirming the identity of the user is that this same identity can be reused when the user's session is reviewed in the process of continuous authentication. The user's identity can be reconfirmed and also be distinguished from an imposter during the consecutive session without being required to interact with the device's authentication protocol. This forms the trade-off between the performance and usability of this multi-step user verification algorithm.


\section{Results and discussion}

For this discussion, two activities from each category i.e. hand-oriented tasks, non-hand oriented tasks and hand-oriented eating tasks have been chosen based on their performance. The user identification model was executed for each of these activities for both benign and adversarial data. The plot for the mean classification probabilities was plotted and a curve to visualise the probability score density was fitted. Figures-2, 3 and 4 show that in the case of adversarial data, the mean probability density drops significantly, from 0.65 to 0.22 in the case of walking (Figure-2), 0.8 to 0.25 in the case of jogging (Figure-2), 0.84 to 0.3 in the case of clapping (Figure-3), 0.8 to 0.3 in the case of typing (Figure-3), 0.85 to 0.3 in the case of drinking (Figure-4) and 0.8 to 0.3 in the case of sandwich (Figure-4). Thus, it was concluded that a different threshold value was needed for each activity (as there's a different model and dataset for each activity). This step of checking that the probability of classification is more than that of the estimated threshold was implemented for both identification as well as authentication. 
\begin{table}[!h]
\caption{Identification accuracy (in \%) from various algorithms}
\label{table:accuracy}
\centering
\begin{tabular}{p{0.2cm}p{0.35cm}p{0.35cm}p{0.35cm}p{0.35cm}p{0.35cm}p{0.35cm}p{0.35cm}p{0.35cm}p{0.35cm}}
\hline
Model &  & \begin{tabular}[c]{@{}c@{}}Random \\ Forest\end{tabular} &  &  & SVM &  &  & \begin{tabular}[c]{@{}c@{}}XG\\ Boost\end{tabular} &  \\ \hline
Act & \begin{tabular}[c]{@{}c@{}}Ph\\ accel\end{tabular} & \begin{tabular}[c]{@{}c@{}}All\\ accel\end{tabular} & \begin{tabular}[c]{@{}c@{}}All\\ sensor\end{tabular} & \begin{tabular}[c]{@{}c@{}}Ph\\ accel\end{tabular} & \begin{tabular}[c]{@{}c@{}}All\\ accel\end{tabular} & \begin{tabular}[c]{@{}c@{}}All\\ sensor\end{tabular} & \begin{tabular}[c]{@{}c@{}}Ph\\ accel\end{tabular} & \begin{tabular}[c]{@{}c@{}}All\\ accel\end{tabular} & \begin{tabular}[c]{@{}c@{}}All\\ sensor\end{tabular} \\ \hline
A & 97.6 & 90.5 & 85.3 & 98.3 & 79.4 & 76.3 & 94.6 & 89.1 & 82.5 \\
B & 96.9 & 93.07 & 86.2 & 96.2 & 75.9 & 74.4 & 94.8 & 86.6 & 82.2 \\
C & 92.3 & 78.6 & 67.9 & 91.9 & 63.7 & 55.8 & 89.1 & 77.6 & 59.8 \\
D & 95.2 & 83.8 & 70.3 & 92.4 & 76 & 57.5 & 93 & 87.9 & 66.4 \\
E & 94.1 & 86.4 & 58.7 & 90.6 & 77.8 & 50 & 91.5 & 84.2 & 57.5 \\
F & 97.8 & 93.3 & 78.4 & 94.5 & 87.5 & 62.8 & 96.2 & 88.7 & 73.5 \\
G & 94.6 & 88.2 & 74.6 & 93.7 & 84.2 & 62.3 & 93.1 & 84.4 & 71.4 \\
H & 96 & 90.5 & 74.8 & 97.2 & 86.5 & 60.6 & 96.8 & 87.4 & 69.1 \\
I & 96 & 82.8 & 66.1 & 94 & 76.4 & 55.2 & 97 & 83.6 & 65.4 \\
J & 96.2 & 83.1 & 68.8 & 96.6 & 77.1 & 54.1 & 95.2 & 83.5 & 66.2 \\
K & 98.8 & 84.9 & 68.2 & 97 & 82.1 & 54.2 & 92.3 & 83.3 & 66.3 \\
L & 96.4 & 83.3 & 65.2 & 93.5 & 77.3 & 53.4 & 94.8 & 84.5 & 66.3 \\
M & 94.5 & 75.3 & 62.7 & 93.8 & 74.7 & 47.1 & 88.2 & 76.5 & 56.3 \\
O & 94.7 & 85.9 & 75.5 & 94.1 & 83.1 & 59.2 & 94.1 & 82.8 & 73.3 \\
P & 94 & 88.7 & 76.2 & 95.1 & 86.4 & 63.9 & 93.7 & 81.3 & 73.9 \\
Q & 95.2 & 91 & 78.2 & 96 & 85.1 & 61.9 & 94.5 & 90 & 70.5 \\
R & 98.8 & 94.7 & 84.1 & 97.6 & 93.2 & 70.6 & 96.2 & 93.6 & 80 \\
S & 95.2 & 80.7 & 67.2 & 91.6 & 75.6 & 48.2 & 91.4 & 79.8 & 58 \\ \hline
Avg & 95.7 & 86.3 & 72.6 & 94.6 & 80.1 & 59.3 & 93.6 & 84.7 & 68.8 \\ \hline
\end{tabular}
\end{table}
\subsection{User Identification}
As described by G. M. Weiss et al \cite{weiss2019smartphone}, to identify a user, for each activity, data (using one or more sensors) for all users is combined. There are fifty-one classes in the identification dataset, and each subject represents a distinct one. In this situation, the training set must include data from all of the participants, and the training and test sets' subjects should overlap. 

The experiment was executed for various sensor combinations with three different algorithms, but as it has already been established by G. M. Weiss et al \cite{weiss2019smartphone}, the sensors from the phone perform better than that from the watch and the accelerometer outperforms the gyrosensor; the rest of this work talks about the results from the phone accelerometer, all accelerometer (phone-accelerometer+watch-accelerometer) and all sensors combined. Out of the three machine learning algorithms used, different algorithms perform better for each activity, but for the average performance overall activities, Random Forest performs the best as it can be inferred from Table-\ref{table:accuracy}.

The activities which perform the best are automatically safer to be used for user verification. Looking at non-hand oriented activities, walking and jogging show the best results for all three algorithms with 97.6 and 96.9 being the highest accuracy that is achieved with RandomForest. Walking and jogging which are means of Gait-based biometrics have also proven to perform well in literature \cite{sun2007research}, \cite{nickel2011benchmarking}. In the instance of considering hand-oriented activities, in the general category, typing and clapping produce the best accuracy with the highest being 97.8 and 98.8 using the RandomForest model. These activities can be recorded using very short periods and thus make a handy case for behavioural biometrics. The hand-oriented eating activities that perform the best are drinking and eating a sandwich with their best case accuracy being 98.8 and 96.4. If we compare the overall results from all activities, clapping and drinking are the best performers. They work better than gait-based activities which have been thoroughly studied for behavioural biometrics. This aspect of our results presents a motivation for implementing motion-based biometrics using hand-oriented activities.
 
\begin{figure*}[!htbp]
        \centering
        \includegraphics[width=12cm,height=7.5cm,keepaspectratio]{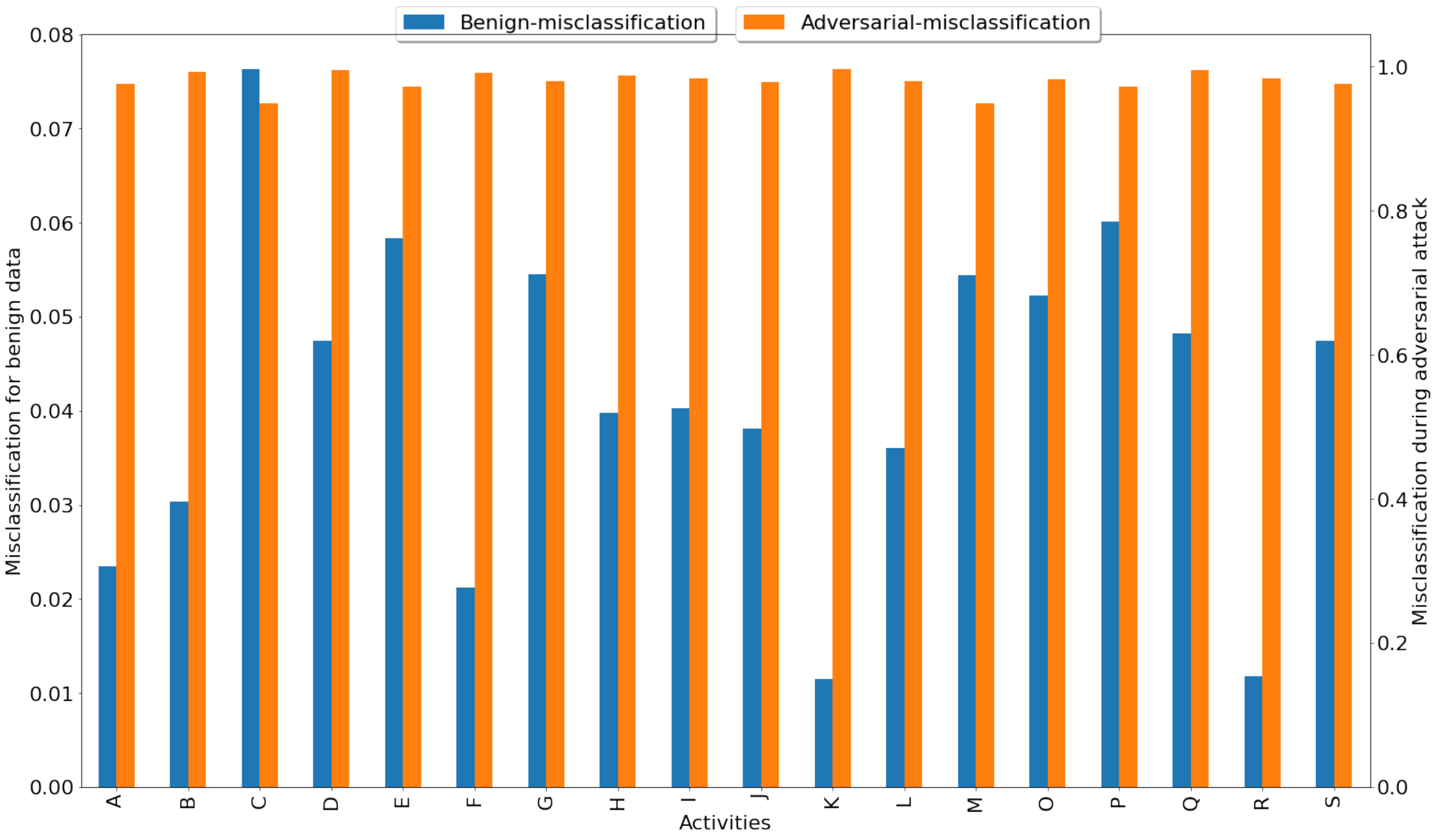}
        \caption{Decline in accuracy \& increase in misclassification due to adversarial attack}
        \label{fig:my_label_db}
\end{figure*}

Although all these accuracies are recorded in the event of benign test samples, Figure-\ref{fig:my_label_db} shows how this model can be fooled into misclassifying users using adversarial data. For our experiment, the adversarial data was created using the Zero Order Optimisation (ZOO) attack. This attack was chosen because it requires no model information and works on tabular as well as time-series data.  Figure-\ref{fig:my_label_db} depicts the difference in the model's misclassification error when benign and adversarial data are used. The y-axis on the left reflects the range of misclassification in benign data (0-0.08), with a maximum error of 0.07, whereas the y-axis on the right shows the range of inaccuracies in adversarial data (0-1), with a maximum misclassification of 0.996. This happens throughout every activity. The accuracy of the model dropped to a low of 0.28 and the samples were highly misclassified. As shown in the mean probability score graphs in Figures-2, 3 and 4, the model does not misclassify with a high confidence score and hence, by using this threshold factor, these misclassifications can be blocked. In our experiment using data from the phone-accelerometer, out of where the test data contained 457 samples, only two samples were misclassified with a probability score that satisfies the threshold value and when employing all sensor data (combining all four sensors), which had 1033 test cases, only 8 were misclassified with a probability score greater than the threshold value. Therefore, we can say that our model is significantly prone to this kind of adversarial attack.
\begin{figure*}[!htbp]
        \centering
        \includegraphics[width=12cm,height=7.5cm,keepaspectratio]{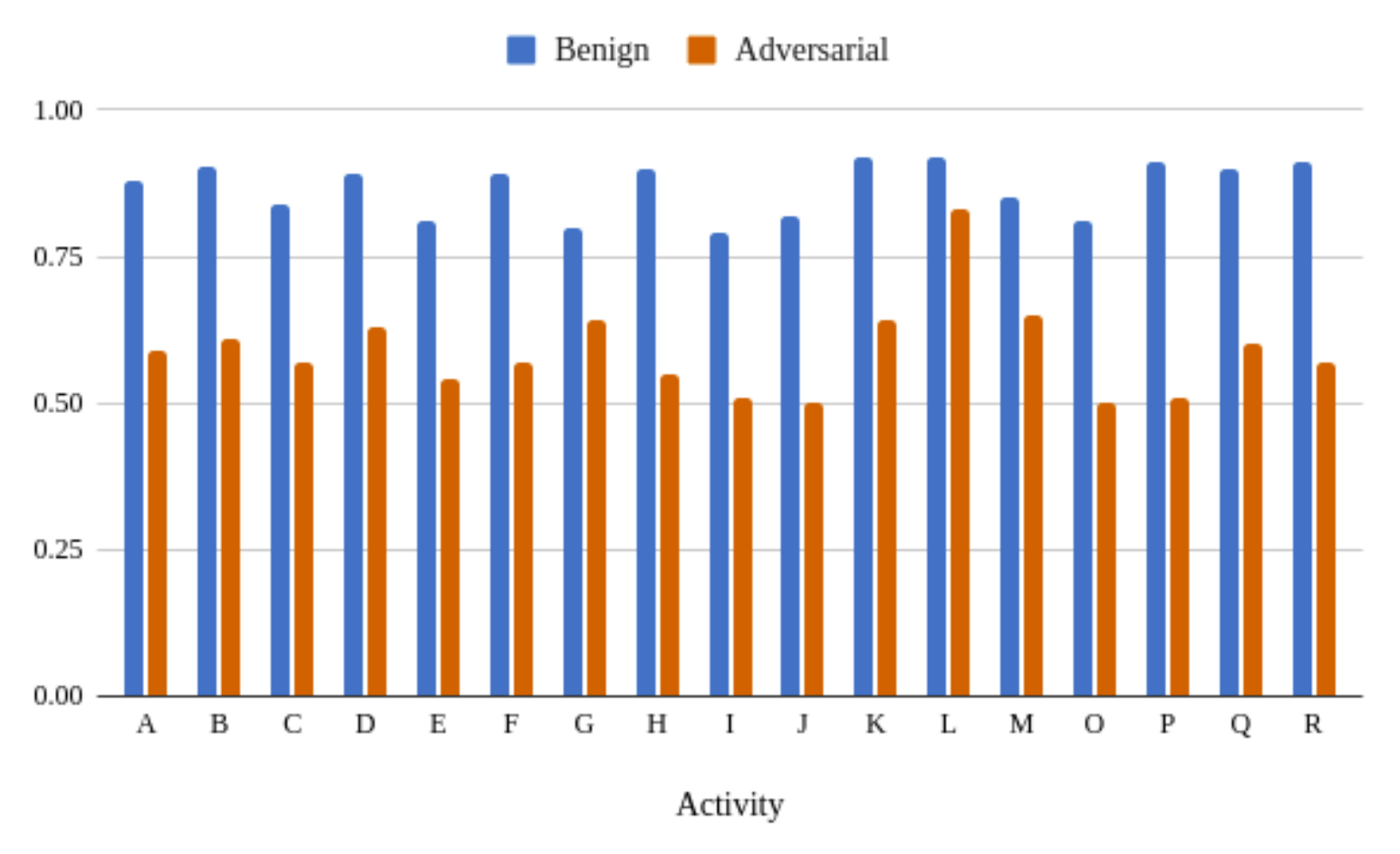}
        \caption{Mean prediction probabilities for Authentication}
        \label{fig:my_label_au}
\end{figure*} 
\subsection{User Authentication}
The results were slightly different for the user authentication experiment which forms the second part of our user verification algorithm. Since the authentication model was different for each user and each activity, the results for different users were averaged with respect to the activities that have been chosen for this discussion. In the case of authentication, the machine learning model was more confident with the misclassifications as compared to that of identification, where the mean probability scores in the case of adversarial misclassification went up to a maximum of 0.35, whereas the probability with which the machine learning model misclassified in the authentication experiment was in the range of (0.50, 0.75), Figure-\ref{fig:my_label_au}. Although there was still a significant amount of difference in the confidence of the model for original data versus the adversarial data, this does demonstrate that the threshold value must be analysed for each model.
\begin{figure*}[!htbp]
        \centering
        \includegraphics[width=12cm,height=7.5cm,keepaspectratio]{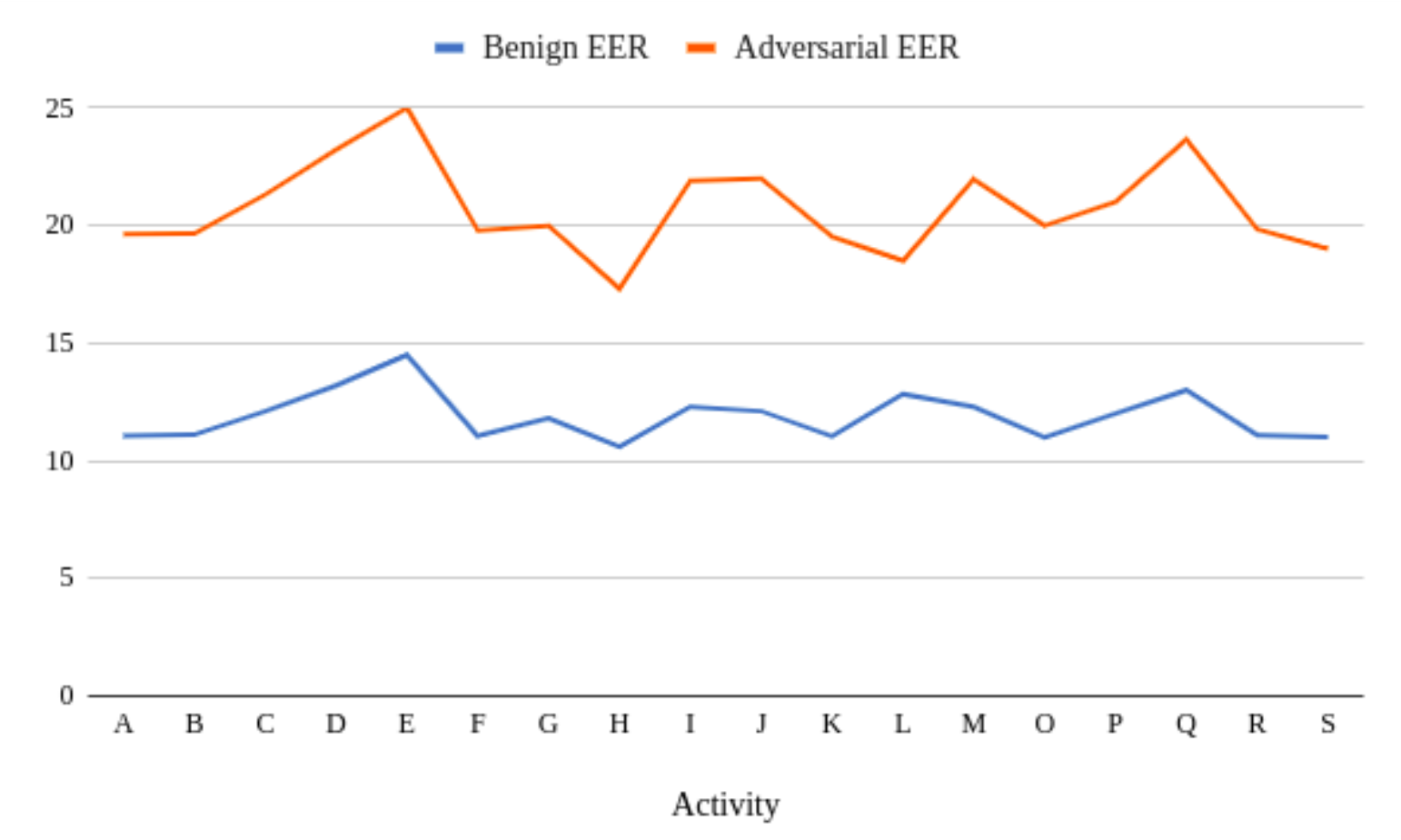}
        \caption{EER for authentication: benign vs adversarial data}
        \label{fig:my_label_eer}
\end{figure*} 
Looking at the Equal Error Rate (EER) for authentication (Figure-\ref{fig:my_label_eer}), when the model was fed with adversarial data, the value of EER kept increasing across all activities, however, the high performing activities from the user identification and user authentication remained similar and the six activities (i.e. walking, jogging, typing, clapping, drinking and eating a sandwich) chosen for this discussion overall performed better. One of the outliers observed in the trend of EER values was the performance of the machine learning model in the case of eating a sandwich where the model performs better with respect to the scores of other activities as well as with respect to the performance of the model in the case of original data. Therefore the relative difference between the two curves of EER is the minimum for the case of eating a sandwich. It can also be observed that the course of the plot remains the same for the rest of the activities as well as for both user identification and user authentication experiments, which reinforce that the proposed algorithm is a stable choice for implementation. The other difference that can be observed from authentication results is that the overall percentage drop in the performance of the model, when tested with adversarial samples, is lower as compared to the identification attempt. Thus the model is appropriately differentiating a user from their imposter and with the threshold value, the results can be better trusted.
As in the case of user identification, the model here also produces a comparably low accuracy in the case of clapping, but the frequency of the activity in the dataset is also an important factor as compared to that of drinking, or other gait based activities such as walking or jogging, which are usually found to occur more than clapping, or specific hand-oriented activities. Accordingly, the low accuracy in the state of clapping or the wide gap between the model's confidence scores for eating a sandwich should not be considered a limitation for this approach, nevertheless, adding more data and better tuning of the model or using a deep learning model with additional data can significantly improve these holes.


\section{CONCLUSION and Future work}

In this work, we proposed a multi-step behavioural biometrics-based user verification algorithm and analysed it for its vulnerability towards adversarial attacks. We also formulated a strategy for defending our algorithm from these kinds of adversarial attacks and presented an empirical evaluation of its effectiveness. Based on the results, we can say that the accelerometers in our phone can readily be used for motion-based biometrics and one way of implementing it is the proposed user verification algorithm which can be easily paired with multi-factor authentication. We employed adversarial attacks to examine the potential risk of attacks on the data which can mislead the model into trusting an imposter. Our plan to defend against the adversarial attacks shows how the duped data does not surpass the user verification model and therefore improves its reliability. The proposed algorithm is multi-step as it not only identifies a user but also ensures that it can be differentiated from an impersonater and hence is suitable for validating a user.

In terms of future work, in addition to integrating this algorithm as a part of multi-factor authentication, there is more work required in improving the accuracy of other sensor combinations, specifically of the smartwatch based sensors. Since human behaviours evolve over time, it is necessary that the user verification model also adapts to the behavioural changes, thus the overall improvement of the model happens with time which will, in turn, increase the security and usability of this approach.


\bibliographystyle{IEEEtran}

\bibliography{references}

\end{document}